\documentclass[runningheads]{llncs}
\usepackage[T1]{fontenc}
\usepackage{graphicx,verbatim}
\usepackage{amsmath, amssymb, amsfonts}
\usepackage{algorithmicx, algorithm, algpseudocode}
\usepackage{hyperref}
\usepackage{xcolor, colortbl, booktabs, multirow, adjustbox, threeparttable}
\usepackage{tikz}
\usepackage{pgfplots}
\newcommand{\tableentry}[1]{\begin{tabular}[l]{@{}l@{}}#1\end{tabular}}
\pgfplotsset{compat=1.18}

\begin{document}
\title{Time-Contrastive Pretraining for In-Context Image and Video Segmentation}

\author{Assefa Wahd\index{Last Name, First Name} \and Jacob Jaremko \and Abhilash Hareendranathan}

\authorrunning{Wahd et al.}
\institute{Department of Radiology and Diagnostic Imaging \\ University of Alberta \\
    \email{\{wahd, jjaremko, hareendr\}@ualberta.ca}}

\maketitle             

\begin{abstract}
In-context learning (ICL) enables generalization to new tasks with minimal labeled data. However, mainstream ICL approaches rely on a gridding strategy, which lacks the flexibility required for vision applications. We introduce \textbf{Temporal}, a time-contrastive self-supervised objective that pretrains a prompt retriever for visual ICL, and formulate ICL as a video object segmentation (VOS) task. Temporal addresses key limitations of grid-based methods that restrict the number and resolution of context images. By reframing ICL as a VOS problem, our approach supports a variable number of context images while preserving their full resolution. To address the challenge of selecting optimal context sets for queries, we pretrain a prompt retriever on videos via self-supervised learning, where adjacent frames serve as positives and distant frames as negatives. For image segmentation, the prompt retriever selects relevant sequences that, when combined with the query, form coherent videos for VOS processing. For video segmentation, it identifies keyframes, predicts their masks using our ICL pipeline, and propagates them throughout the sequence. When evaluated on MICCAI FLARE 2022, our method achieves substantial improvements over baselines: 90.95\% Dice score for image segmentation (10.64\% improvement) and 92.45\% Dice for video segmentation (14.88\% improvement).

\keywords{Time-Contrastive Learning \and In-Context Learning \and Video Object Segmentation}
\end{abstract}

\section{Introduction}
\begin{figure}[!t]
    \centering
    \includegraphics[width=\textwidth]{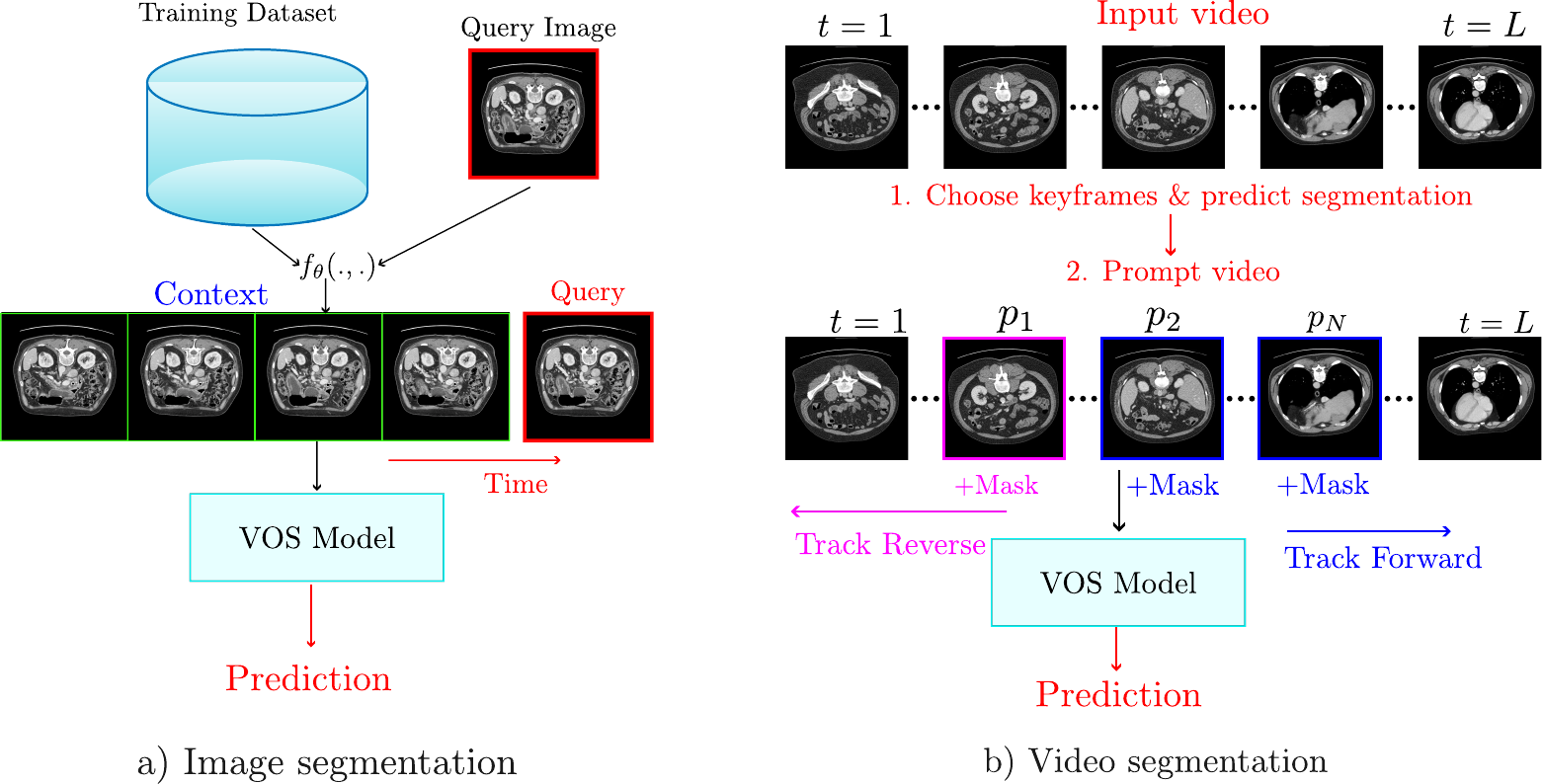}
      \caption{\textbf{Model architecture.} (a) \textit{Visual in-context learning via VOS:} The retriever $f_\theta$ selects the optimal context set $P$ for a query image $x_q$. The query image $x_q$ is appended to the context to create a video sequence, and the VOS model predicts the segmentation mask for $x_q$ using this sequence. (b) \textit{Video segmentation with ICL:} 1) Select keyframes (Section \ref{sec:video_segmentation}), 2) Predict masks for keyframes using step (a), 3) Propagate masks across the video.}
    \label{fig:arch}
  \end{figure}

Medical imaging generates vast quantities of unlabeled data, yet acquiring dense annotations remains costly. This challenge has motivated the development of approaches requiring fewer labels, such as \emph{self-supervised learning} \cite{balestrieroCookbookSelfSupervisedLearning2023,chenSimpleFrameworkContrastive2020,oquabDINOv2LearningRobust2024,heMaskedAutoencodersAre2021,heMomentumContrastUnsupervised2020,assranSelfSupervisedLearningImages2023}, \emph{semi-supervised learning} \cite{vanengelenSurveySemisupervisedLearning2020}, and \emph{transfer learning} \cite{weissSurveyTransferLearning2016}. Recently, foundation models have introduced new paradigms such as \emph{visual prompting} and \emph{in-context learning} (ICL) \cite{yuVisualTuning2024,dennerVisualPromptEngineering2024,fosterFlexibleVisualPrompts2023,barVisualPromptingImage,zhangInstructMeMore2023}, enabling adaptation to new tasks with minimal or no fine-tuning.

In-context learning (ICL)—where models perform new tasks by conditioning on input-output examples without parameter updates—has been remarkably successful in natural language processing \cite{brownLanguageModelsAre2020}. It offers a compelling vision for medical imaging: a single model could generalize across diverse segmentation tasks by "reasoning" over relevant annotated examples. However, adapting ICL to computer vision presents unique challenges. We highlight two critical issues: (1) selecting the most relevant context images, and (2) effectively modeling interactions between the context set and query images. An ideal visual ICL model must support a variable number of context images, and presever their full resolution. In textual tasks, input-output pairs can be trivially concatenated to form prompts; for visual ICL, however, it is less clear how to handle a variable number of images.

 Early attempts at visual ICL rely on a \textit{gridding} strategy, in which input-output image pairs (i.e., the in-context examples) are arranged in a grid-like format alongside a query image \cite{barVisualPromptingImage,zhangWhatMakesGood2023,zhangInstructMeMore2023,wangImagesSpeakImages2023a}. However, this strategy limits the number of context images and necessitates downsampling, reducing image detail—a critical limitation for medical applications where fine details matter. More recent works have focused on context selection methods. Zhang et al. \cite{zhangWhatMakesGood2023} found that selecting relevant context examples significantly impacts performance and proposed both unsupervised (CLIP-based \cite{radfordLearningTransferableVisual2021}) and supervised retrieval techniques. Subsequent innovations included padding images with learnable vectors \cite{zhangInstructMeMore2023} and combining channel and spatial features when computing similarity \cite{sunExploringEffectiveFactors2023}.
  
A promising alternative emerged when Foster et al. \cite{fosterFlexibleVisualPrompts2023} framed ICL as a video object segmentation (VOS) task. In this approach, context images are treated as frames in a synthetic video, with the VOS model \cite{chengXMemLongTermVideo2022} segmenting the final frame (the query image). This allows handling full-resolution images and variable-sized context sets. However, they relied on pre-trained CLIP embeddings for context retrieval, which may not be optimal for domain-specific medical imaging tasks. Our work, named \textbf{Temporal}, extends this VOS-based framework with two key innovations: (1) a self-supervised time-contrastive approach to context retrieval and (2) a unified framework for both image and video segmentation. Our multipositive time-contrastive pretraining builds on Time-Contrastive Networks (TCN) \cite{sermanetTimeContrastiveNetworksSelfSupervised2018}, enhanced with multipositive cross-entropy loss instead of traditional triplet loss. Unlike supervised retrieval methods \cite{zhangWhatMakesGood2023}, we leverage temporal proximity in videos as a natural supervisory signal—temporally adjacent frames likely contain semantically similar information, creating positive pairs without manual labels. 

We provide a comprehensive analysis of context selection strategies, examining size, order, and diversity considerations, and introduce a diversity-aware retrieval mechanism using our time-contrastive network. For video data, we propose an efficient keyframe-based retrieval and propagation approach that significantly outperforms existing methods. To the best of our knowledge, Temporal is the first approach to combine time-contrastive networks with multipositive cross-entropy loss for self-supervised context retrieval from videos. While this paper focuses on using this technique as a prompt retriever for visual in-context learning, future work will explore broader applications of multipositive TCNs as generic SSL networks.  Our method achieves up to 92.45\% Dice on video benchmarks—a 14.88\% improvement over semi-supervised VOS baselines. Figure~\ref{fig:arch} illustrates our approach for both image and video segmentation. You can find the code at \href{https://github.com/aswahd/Temporal}{github.com/aswahd/Temporal}.

\begin{figure}[!t]
    \includegraphics[width=\textwidth]{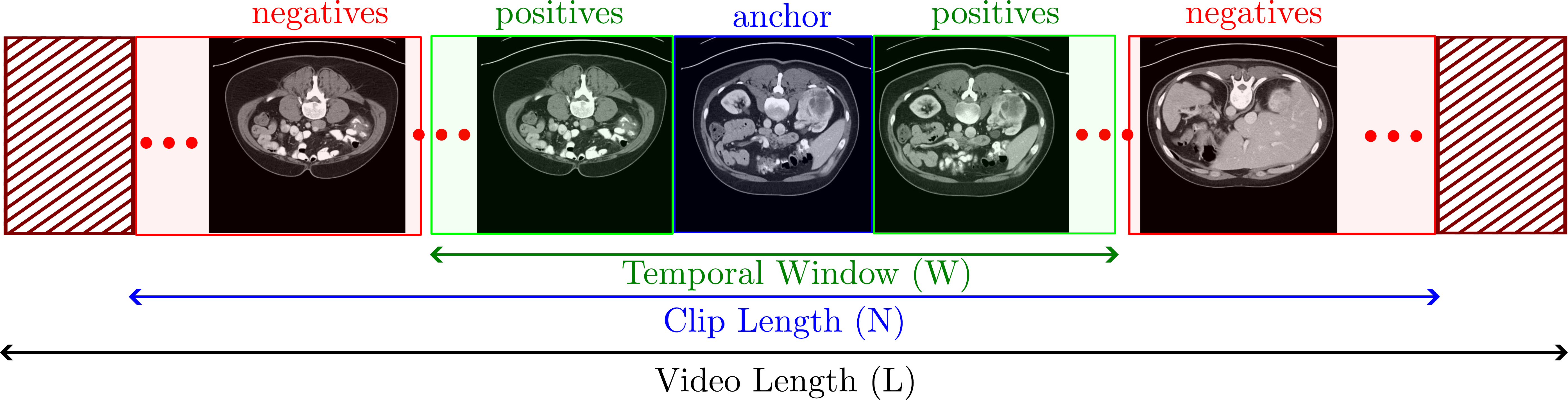}
     \caption{\textit{Multi-positive sampling strategy}. For each anchor image in a clip, we sample $M$ temporally adjacent frames as positives and all other frames in the clip as negatives. Shaded regions are frames that were not sampled in a minibatch.}
     \label{fig:sampling}
 \end{figure}

\section{Method}
\label{sec:method}
Given a video sequence $V = (x_1, x_2, \dots, x_T)$ where $x_t$ represents a frame at time $t$, a VOS model $\mathcal{V}_\phi$ generates segmentation masks $\hat y_t$ by utilizing both current and historical frame information:
\begin{equation}
  \hat y_t = \mathcal{V}_\phi(x_1, y_1, x_2, y_2, \dots, x_{t-1}, y_{t-1}, x_t),
\end{equation}
where $\hat y_t$ is the predicted segmentation for frame $t$.

Our approach builds on SAM2 \cite{raviSAM2Segment} as the foundation VOS model, which incorporates: 1) \emph{ViT image encoder} for embeddings, 2) \emph{Memory Encoder/Bank} to store/update past predictions, 3) \emph{Memory Attention} for historical context fusion, and 4) \emph{Promptable Mask Decoder} integrating embeddings, prompts, and memory for segmentation. We refer readers to \cite{raviSAM2Segment} for a detailed description of SAM2.

\subsection{Visual In-Context Learning}

In-context learning (ICL) infers an output for a query input by conditioning on examples. Given a dataset $\mathcal{D} = \{(x_n, y_n)\}_{n=1}^N$ of input-output pairs and a query $x_q$, we predict $\hat y_q$ using a model $\mathcal{V}_\phi$ conditioned on a context \( P = \{(x_{c_1}, y_{c_1}), \dots, (x_{c_K}, y_{c_K})\} \subseteq \mathcal{D}.  \)

To generalize beyond grid layouts, we treat $P$ and $x_q$ as frames in a \emph{synthetic video}, where each pair $(x_{c_k}, y_{c_k})$ is a labeled frame and $x_q$ is the unlabeled final frame. A VOS model then segments $x_q$:
\begin{equation}
\mathcal{V}_\phi(P, x_q) = \mathcal{V}_\phi\bigl(x_{c_1}, y_{c_1}, x_{c_2}, y_{c_2}, \dots, x_{c_K}, y_{c_K},\, x_q\bigr),
\label{eq:vos_icl}
\end{equation}
where $\mathcal{V}_\phi$ is a pretrained foundation model. A prompt retriever $f_\theta$ selects the optimal context as outlined in the next section.

\subsection{Multipositive Time-Contrastive Pretraining}
Our objective is to sample multiple positive examples along with a substantial number of negatives for each anchor frame in a video, and to train the network using a contrastive loss. We formalize this as follows:

\noindent\emph{Sampling Strategy:} For each video \( V \) of length \( L \), we sample a contiguous clip \( C = \{ t_{\text{start}}, t_{\text{start}} + 1, \dots, t_{\text{start}} + N - 1 \} \), where \( N \) denotes the clip length and \( t_{\text{start}} \sim \mathcal{U}(1, L - N) \). All frames in the clip are used in a minibatch.

\noindent\emph{Generating Label Matrix:} For each anchor frame \( i \in C \), positive pairs are defined as frames within a temporal window \( \mathcal{W}_i = [\max(1, i - \lfloor W/2 \rfloor), \min(N, i + \lfloor W/2 \rfloor)] \), where \( W \) specifies the window size. We sample $M$ frames from the window.  This generates a binary label matrix \( \mathbf{A} \in \{0,1\}^{N \times N} \):

\begin{equation}
\mathbf{A}_{i,j} = \begin{cases} 
1 & \text{if } |i - j| \leq \lfloor W/2 \rfloor \\
0 & \text{otherwise}
\end{cases}
\end{equation}

To prevent trivial self-similarity matches, diagonal elements \( \mathbf{A}_{i,i} \) are masked to 0.  

\noindent\emph{Multi-View Augmentation:} We generate two augmented views \( \tilde{C}_1 \) and \( \tilde{C}_2 \) through random geometric (e.g., translation) and photometric (e.g., Gaussian blurring) transformations. The label matrix expands to encode multi-view relationships:

\begin{equation}
\mathbf{\hat{A}} = \begin{bmatrix}
\mathbf{A} & \mathbf{A} + \mathbf{I}_N \\
\mathbf{A} + \mathbf{I}_N & \mathbf{A}
\end{bmatrix}
\end{equation}

where \( \mathbf{I}_N \) denotes the \( N \times N \) identity matrix. This formulation ensures that:
\begin{itemize}
    \item Temporal neighbors within each view remain positives
    \item Temporal neighbors across views are positives
    \item Two views of the same frame are positives   (i.e., $+\mathbf{I}_N$).
\end{itemize}
For minibatch training, we sample \( B \)  video clips, and extend the above label matrix to consider frames from different videos as negatives. The final label matrix is a block diagonal matrix with $B$ blocks. See Fig. \ref{fig:arch} for the overall architecture and \ref{fig:sampling} for an illustration of a sampled minibatch.

\noindent\emph{Multipositive Time-Contrastive Loss:} Each anchor, in a minibatch  of shape $(B \times N, C, H, W)$, has $2M-1$ positives and $2B(N-M)$ negatives. The encoder $f_\theta$ projects the input batch into a set of embeddings $\mathbf{z} = \{z_1, z_2, \dots, z_{2BN}\}$. Each embedding is L2 normalized to obtain $\hat{z}_i = z_i/\|z_i\|$. The multipositive contrastive loss is defined as:
\begin{equation}
  \mathcal{L}_{\text{contrastive}} = -\frac{1}{2BN} \sum_{i=1}^{2BN} \frac{1}{|P(i)|} \sum_{j \in P(i)} \log \frac{\exp\!\left(\frac{\hat{z}_i^\top \hat{z}_j}{\tau}\right)}{\exp\!\left(\frac{\hat{z}_i^\top \hat{z}_j}{\tau}\right) + \sum_{k \in N(i)} \exp\!\left(\frac{\hat{z}_k^\top \hat{z}_j}{\tau}\right)},
  \label{eqn:nt-xent}
\end{equation}
where $P(i)$ denotes the set of positive samples for the $i$-th anchor, $N(i)$ represents the corresponding set of negatives, and $\tau$ is the temperature parameter.

\subsection{Inference}
\label{sec:video_segmentation}
For a test image \( x_q \), the inference for images proceeds as follows:
\begin{enumerate}
    \item \emph{Context Retrieval}: Compute the query embedding \( z_q = f_\theta(x_q) \) and retrieve the top-\( K \) most similar images from the training database \( \mathcal{D} \) using cosine similarity between \( z_q \) and stored embeddings.
    \item \emph{Synthetic Video Construction}: Form a temporal sequence \( V = \{x_{c_1}, \dots, x_{c_K}, x_q\} \) where \( \{x_{c_i}\} \) are retrieved contexts and \( x_q \) is appended as the final "frame".
    \item \emph{Mask Prediction}: Process \( V \) through the VOS model \( \mathcal{V}_\phi \) to generate the query segmentation \( \hat{y}_q \) via Eq.~\ref{eq:vos_icl}. Fig. \ref{fig:arch}(b) illustrates this pipeline.
\end{enumerate}

\noindent\textbf{Video Segmentation:} For video input \( V = \{x_1, \dots, x_L\} \), we generate segmentations \(\{\hat{y}_1, \dots, \hat{y}_L\}\) through a four-step process:

\begin{enumerate}
    \item \emph{Similarity Computation}: We embed each frame \(x_t\) as \(z_t = f_\theta(x_t)\) and compute its similarity to the training database \(\mathcal{D}\) as \(s_t = \max_{d \in \mathcal{D}} \frac{z_t \cdot d}{\|z_t\|\|d\|}\). The top-\(K\) frames with highest similarity scores form candidate set \(\mathcal{S}\).
    
    \item \emph{Diverse Keyframe Selection}: From \(\mathcal{S}\), we select \(Q\) temporally diverse keyframes to form \(\mathcal{Q}\) by ensuring minimum temporal distance \(\min_{x_j \in \mathcal{Q}} |t_i - t_j| > \tau\) between selected frames, prioritizing frames with higher similarity scores.
    
    \item \emph{Confidence-Based Filtering}: For each keyframe \(x_q \in \mathcal{Q}\), we apply our Temporal model to predict both segmentation mask and confidence score: \((\hat{y}_q, c_q) = \text{Temporal}(x_q, \mathcal{D})\). We retain only high-confidence predictions as prompts: \(P = \{(x_q, \hat{y}_q) \,|\, c_q \geq \gamma\}\).
    
    \item \emph{Bidirectional Propagation}: Since keyframes may occur anywhere in the video, we feed prompts \(P\) into the VOS model \(\mathcal{V}_\phi\) and propagate masks bidirectionally (forward and backward) to generate segmentations for all frames: \(\{\hat{y}_1, \hat{y}_2, \dots, \hat{y}_L\} = \mathcal{V}_\phi(V,P)\) (Fig.~\ref{fig:arch} b)).
\end{enumerate}

\noindent\textbf{Fine-tuning.} To bridge the domain gap between natural and medical images, we fine-tune the underlying VOS model using "synthetic" videos constructured from 2D images in the training set.  For each image $\mathbf{x}_i$, we: 1) select top-$K$ context images via $f_\theta$, 2) construct video $V_i = \{\mathbf{x}_1, \dots, \mathbf{x}_K, \mathbf{x}_Q\}$ sorted by similarity, and 3) train $\mathcal{V}_\phi$ on random clips $v_i \subset V_i$ to minimize \(\min_\phi \mathcal{L}_{seg}(\mathcal{V}_\phi(v_i), y)\).  Fine-tuning significantly improves both image and video segmentation as shown in Tables~\ref{tab:main_res} \& \ref{tab:exp_vos}.

\section{Experiments}
\label{sec:exp}

\begin{figure}[!t]
  \includegraphics[width=\linewidth]{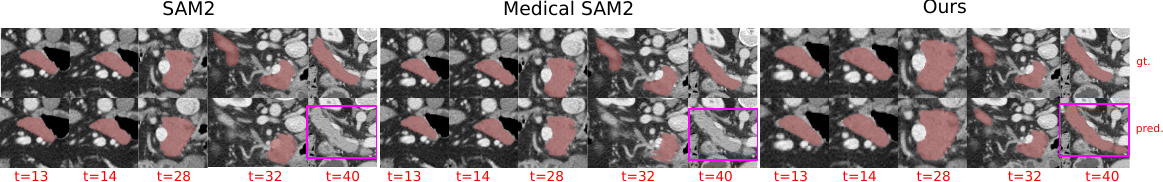}
   \caption{\textit{Qualitative segmentation results} on Pancreas. Our approach (right) produces more accurate organ segmentations and tracking compared to baselines. The baselines fail to track the organ after $t=32$.}
   \label{fig:qualitative_results}
\end{figure}

\noindent\emph{Representative methods.} We compare against grid-based (Zhang et al.~\cite{zhangWhatMakesGood2023}; CLIP-based retrieval) and VOS-based ICL (Foster et al.~\cite{fosterFlexibleVisualPrompts2023}; CLIP-based retrieval). We adopt SAM-2 as the video foundation model since it represents the current SOTA in generic video segmentation.  

\noindent\emph{Datasets.} We evaluate our approach on the MICCAI FLARE 2022 dataset, a benchmark dataset for CT scan segmentation. The dataset includes both annotated scans and a substantial collection of unlabeled data, which we utilize for self-supervised pretraining. Specifically, we use 40 scans to construct the context set and 10 scans for testing. These 10 scans comprise 922 slices containing foreground annotations.

\noindent\emph{Implementation Details.} Our time-contrastive pretraining employs a ResNet-50 \cite{heDeepResidualLearning2015} backbone. We sample 16-frame video clips with a stride of 2 and apply random augmentations (e.g., rotation, horizontal flipping, and Gaussian blurring). For the cross entropy loss, the temperature is set to $\tau=0.1$, and two adjacent frames are selected as positives per anchor. Training runs for 500 epochs with a minibatch of 4 clips (resulting in a batch size of $16 \times 2 \times 4 = 128$ images) using the AdamW optimizer \cite{loshchilovDecoupledWeightDecay2019} (learning rate $lr=10^{-3}$ and weight decay $10^{-4}$). The hyperparameters were chosen via grid search over $lr \in \{10^{-4},10^{-3},10^{-2}\}$ and $\tau \in \{0.1,0.5,1.0\}$. The VOS model utilizes a SAM2-B+ backbone \cite{raviSAM2Segment}. For image segmentation, we empirically found that 10 context images work best (see Table~\ref{tab:main_res}). For video segmentation, we sample 20 keyframes. Fine-tuning is performed for 100 epochs with a learning rate of $10^{-5}$, a batch size of 1, and the AdamW optimizer.

\subsection{Results}
\label{sec:exp_main}
\emph{Image segmentation:} The grid-based method with unsupervised context retrieval performs poorly on medical images, achieving only around 5\% Dice score. So, we abandon this approach in favor of VOS-based methods. Table~\ref{tab:main_res} shows that Temporal outperforms unsupervised prompt retrieval methods including Dino v2 \cite{oquabDINOv2LearningRobust2024}, CLIP \cite{radfordLearningTransferableVisual2021}, and MedCLIP \cite{wangMedCLIPContrastiveLearning2022}. Our method achieves 83.55\% Dice versus 80.31\% for the best baseline. After fine-tuning, Temporal reaches \textbf{90.95\%} Dice, a \textbf{10.64\%} improvement. Performance gains are particularly significant for challenging organs like Pancreas (79.53\% versus baseline's 49.10\%).

\noindent\emph{Video Segmentation:} We evaluate on 10 abdominal CT videos (average length: 95 frames) comparing two paradigms:  
\begin{itemize}
    \item \textbf{Semi-supervised baselines}: SAM-2 \cite{raviSAM2Segment} and Medical SAM-2 \cite{zhuMedicalSAM22024} require first-frame manual prompts. So, we provide ground-truth masks for the first annotated slice of each organ.
    \item \textbf{Automated VOS}: Temporal automatically identifies the target organs without manual prompts, and tracks them throughout the entire sequence.
\end{itemize}

Table~\ref{tab:exp_vos} highlights our method's superiority: Temporal achieves \textbf{81.93\%} Dice fully automatically compared to SAM2's 77.57\% with manual prompts. After fine-tuning (Section \ref{sec:method}), Temporal achieves \textbf{92.23\%} Dice, outperforming Medical SAM-2 (77.83\%) by \textbf{14.4\%}, and \textbf{92.45\%} in semi-supervised segmentation. The improvement is particularly dramatic on challenging organs like Pancreas, where the best baseline achieves only 35.57\% Dice compared to Temporal's \textbf{82.03\%} after fine-tuning. See Fig.~\ref{fig:qualitative_results} for qualitative results.

\begin{table*}[!t]
  \centering
  \caption{\textbf{Image segmentation} results (Dice, \%) of Temporal and baselines on the FLARE 2022 dataset. "ctx" denotes the number of context images used. All baseline models are evaluated with ctx=10. (RK: Right Kidney, Spleen, Panc: Pancreas, Aorta, IVC: Inferior Vena Cava, Gall: Gallbladder, Eso: Esophagus, Stom: Stomach, LK: Left Kidney). Fine-tuning (ft) improves performance.}
  \label{tab:main_res}
  \resizebox{\linewidth}{!}{%
  \begin{tabular}{l|cccccccccc|c}
  \hline
  \rowcolor{blue!40!white!20}
  Model & Liver & RK & Spleen & Panc. & Aorta & IVC & Gall. & Eso. & Stom. & LK & Avg \\ \hline
  \rowcolor{cyan!40!white!20}
  \multicolumn{12}{c}{\cellcolor{orange!30!yellow!20}\textbf{Grid-based methods}} \\
  \tableentry{Zhang et al.~\cite{zhangWhatMakesGood2023}} (CLIP) & \multicolumn{11}{c}{N/A}\\
  \midrule
  \rowcolor{orange!20!yellow!10}
  \multicolumn{12}{c}{\cellcolor{gray!8}\textbf{VOS-based methods}} \\
  \rowcolor{gray!10} 
    \tableentry{Foster et al.~\cite{fosterFlexibleVisualPrompts2023} (Random, 5 evals)} & \tableentry{0.8999} & \tableentry{0.8904} & \tableentry{0.8561} & \tableentry{0.3006} & \tableentry{0.9278} & \tableentry{0.6998} & \tableentry{0.5378} & \tableentry{0.4127} & \tableentry{0.6349} & \tableentry{0.9003} & 0.7060 \\
  \tableentry{Foster et al.~\cite{fosterFlexibleVisualPrompts2023}  (CLIP)} & 0.9234 & 0.8968 & 0.9136 & 0.4886 & 0.9490 & 0.8252 & 0.6142 & 0.6434 & 0.7291 & 0.8583 & 0.7842 \\
  \rowcolor{gray!10} 
  \tableentry{Foster et al.~\cite{fosterFlexibleVisualPrompts2023} (MedCLIP)} & 0.8950 & 0.8486 & 0.8605 & 0.4464 & 0.9535 & 0.8207 & 0.6647 & 0.6758 & 0.7501 & 0.8097 & 0.7724\\
  \tableentry{Foster et al.~\cite{fosterFlexibleVisualPrompts2023}  (Dino v2)} & \tableentry{0.9161} & \tableentry{0.9003} & \tableentry{0.8815} & \tableentry{0.4910} & \tableentry{0.9485} & \tableentry{0.8432} & \tableentry{0.7519} & \tableentry{0.6730} & \tableentry{0.7509} & \tableentry{0.8747} & 0.8031 \\
  \midrule
  \multicolumn{12}{c}{\cellcolor{gray!8}\textbf{Ours}} \\
  \rowcolor{green!10} 
  Temporal (ctx=5) & 0.9088 & 0.9003 & 0.9222 & 0.6105 & 0.9526 & 0.8185 & 0.7579 & 0.7096 & 0.7194 & 0.8881 & 0.8188 \\ 
  Temporal (ctx=10) & 0.9243 & 0.9105 & 0.9221 & 0.6211 & 0.9560 & 0.8364 & 0.8130 & 0.7228 & 0.7499 & 0.8991 & 0.8355 \\ 
  \rowcolor{green!10} 
  \tableentry{Temporal (ctx=5,ft)} & 0.9504 & 0.9115 & 0.9578 & 0.7872 & 0.9669 & 0.9204 & 0.8361 & 0.8619 & 0.8585 & 0.8980 & 0.8949 \\ 
  \tableentry{Temporal (ctx=10, ft, \textcolor{red}{topk})} & 0.9506 & 0.9032 & 0.9371 & 0.7860 & 0.9669 & 0.9214 & 0.8598 & 0.8772 & 0.8671 & 0.8958 & 0.8965 \\
  \rowcolor{green!10} 
  \tableentry{Temporal (ctx=10,ft)} & 0.9631 & 0.9202 & 0.9650 & 0.7953 & 0.9671 & 0.9218 & 0.9073 & 0.8697 & 0.8764 & 0.9090 & 0.9095 \\
  \hline
  \end{tabular}}
\end{table*}

\begin{table*}[!t]
  \centering
  \caption{\textbf{Video segmentation} results (Dice, \%). SAM2 and Medical SAM2 are evaluated in a semi-supervised setting as they require prompts to propagate. Temporal can automatically select context without supervision and achieves superior performance.}
  \label{tab:exp_vos}
  \resizebox{\linewidth}{!}{%
  \begin{tabular}{l|cccccccccc|c}
  \hline
  \rowcolor{blue!40!white!20}
  Method & Liver & RK & Spleen & Panc. & Aorta & IVC & Gall. & Eso. & Stom. & LK & Avg \\ \hline
  \rowcolor{cyan!40!white!20}
  \multicolumn{12}{c}{\cellcolor{orange!30!yellow!20}\textbf{Baselines (Semi-supervised)}} \\
  SAM-2 (mask prompt) & 0.9121 & 0.9638 & 0.8845 & 0.3557 & 0.9497 & 0.8021 & 0.8737 & 0.6497 & 0.4395 & 0.9257 & 0.7757 \\ 
  \tableentry{Medical SAM-2 (mask prompt)} & 0.9070 & 0.8991 & 0.9650 & 0.3402 & 0.8932 & 0.7828 & 0.7238 & 0.6970 & 0.6786 & 0.8358 & 0.7723 \\
  \midrule
  \rowcolor{orange!20!yellow!10}
  \multicolumn{12}{c}{\cellcolor{gray!8}\textbf{Ours}} \\
  Temporal (ctx=10) & 0.9263 & 0.8867 & 0.9280 & 0.5970 & 0.9526 & 0.8633 & 0.8105 & 0.6281 & 0.7149 & 0.8853 & 0.8193\\
  \rowcolor{green!10}
  \tableentry{Temporal \\(ctx=10, fine-tuned)} & \textbf{0.9739} & \textbf{0.9625} & \textbf{0.9532}  & \textbf{0.8132} & \textbf{0.9669} & \textbf{0.9219} & \textbf{0.8928} & \textbf{0.8686} & \textbf{0.9118} & \textbf{0.9581} & \textbf{0.9223} \\
  \tableentry{Temporal \\(fine-tuned, mask prompt)} & \textbf{0.9765} & \textbf{0.9625} & \textbf{0.9837}  & \textbf{0.8203} & \textbf{0.9690} & \textbf{0.9351} & \textbf{0.8927} & \textbf{0.8875} & \textbf{0.8586} & \textbf{0.9586} & \textbf{0.9245} \\
  \hline
  \end{tabular}}
\end{table*}

\paragraph{Impact of Context Size.} To evaluate the effect of the number of annotated context images, we varied the number of context examples in our experiments. As shown in Table \ref{tab:main_res}, increasing the number of context images results in improved performance: for example, 74.70\% Dice with 5 context images versus 81.83\% Dice with 10.

\paragraph{Diversity-Aware Context Selection.} We compare standard top-$K$ context retrieval with a diversity-aware selection strategy. Our two-stage approach: (1) retrieves top-$K$ most similar training images to the query using cosine similarity on SSL features, then (2) greedily selects top-$Q$ diverse candidates.  The core idea is to minimize redundancy in the context set, ensuring each selected example contributes unique information. For each candidate $\mathbf{z}_j$ in the $K$-subset, we iteratively select the sample maximizing:
\[
\text{score}(z_j) = \text{sim}(z_j, z_{\text{query}}) - \lambda \cdot \max_{\mathbf{z}_k \in \mathcal{S}} \text{sim}(z_j, z_k),
\]
where $\mathcal{S}$ contains already selected samples and $\lambda$ balances relevance versus diversity. In our experiments, $\lambda=0.7$ was determined through validation. The results in Table \ref{tab:main_res} show that diverse sampling improves Dice by $\uparrow 2.1\%$ (context size 5) and $\uparrow 1.30\%$ (size 10) over top-$K$ selection, highlighting the importance of context diversity.

\section{Conclusion}
\label{sec:conclusion}
We introduced Temporal, a self-supervised learning objective for pretraining a prompt retriever, and we formulated visual in-context learning as a video object segmentation. When evaluated on the MICCAI FLARE 2022 dataset, our approach demonstrates substantial improvements, achieving a 9.23\% and 14.88\% increase in Dice scores in image and video segmentation, respectively.

\bibliographystyle{splncs04}
\bibliography{refs}

\begin{thebibliography}{10}
\providecommand{\url}[1]{\texttt{#1}}
\providecommand{\urlprefix}{URL }
\providecommand{\doi}[1]{https://doi.org/#1}

\bibitem{assranSelfSupervisedLearningImages2023}
Assran, M., Duval, Q., Misra, I., Bojanowski, P., Vincent, P., Rabbat, M., LeCun, Y., Ballas, N.: Self-{{Supervised Learning}} from {{Images}} with a {{Joint-Embedding Predictive Architecture}} (Apr 2023). \doi{10.48550/arXiv.2301.08243}

\bibitem{balestrieroCookbookSelfSupervisedLearning2023}
Balestriero, R., Ibrahim, M., Sobal, V., Morcos, A., Shekhar, S., Goldstein, T., Bordes, F., Bardes, A., Mialon, G., Tian, Y., Schwarzschild, A., Wilson, A.G., Geiping, J., Garrido, Q., Fernandez, P., Bar, A., Pirsiavash, H., LeCun, Y., Goldblum, M.: A {{Cookbook}} of {{Self-Supervised Learning}} (Jun 2023). \doi{10.48550/arXiv.2304.12210}

\bibitem{barVisualPromptingImage}
Bar, A., Gandelsman, Y., Darrell, T., Globerson, A., Efros, A.A.: Visual {{Prompting}} via {{Image Inpainting}}

\bibitem{brownLanguageModelsAre2020}
Brown, T.B., Mann, B., Ryder, N., Subbiah, M., Kaplan, J., Dhariwal, P., Neelakantan, A., Shyam, P., Sastry, G., Askell, A., Agarwal, S., {Herbert-Voss}, A., Krueger, G., Henighan, T., Child, R., Ramesh, A., Ziegler, D.M., Wu, J., Winter, C., Hesse, C., Chen, M., Sigler, E., Litwin, M., Gray, S., Chess, B., Clark, J., Berner, C., McCandlish, S., Radford, A., Sutskever, I., Amodei, D.: Language {{Models}} are {{Few-Shot Learners}} (Jul 2020). \doi{10.48550/arXiv.2005.14165}

\bibitem{chenSimpleFrameworkContrastive2020}
Chen, T., Kornblith, S., Norouzi, M., Hinton, G.: A {{Simple Framework}} for {{Contrastive Learning}} of {{Visual Representations}} (Jun 2020). \doi{10.48550/arXiv.2002.05709}

\bibitem{chengXMemLongTermVideo2022}
Cheng, H.K., Schwing, A.G.: {{XMem}}: {{Long-Term Video Object Segmentation}} with an {{Atkinson-Shiffrin Memory Model}} (Jul 2022). \doi{10.48550/arXiv.2207.07115}

\bibitem{dennerVisualPromptEngineering2024}
Denner, S., Bujotzek, M., Bounias, D., Zimmerer, D., Stock, R., Jäger, P.F., {Maier-Hein}, K.: Visual {{Prompt Engineering}} for {{Medical Vision Language Models}} in {{Radiology}} (Aug 2024)

\bibitem{fosterFlexibleVisualPrompts2023}
Foster, T., Croitoru, I., Dorfman, R., Edlund, C., Varsavsky, T., Almazán, J.: Flexible visual prompts for in-context learning in computer vision (Dec 2023). \doi{10.48550/arXiv.2312.06592}

\bibitem{heMaskedAutoencodersAre2021}
He, K., Chen, X., Xie, S., Li, Y., Dollár, P., Girshick, R.: Masked {{Autoencoders Are Scalable Vision Learners}}. https://arxiv.org/abs/2111.06377v3 (Nov 2021)

\bibitem{heMomentumContrastUnsupervised2020}
He, K., Fan, H., Wu, Y., Xie, S., Girshick, R.: Momentum {{Contrast}} for {{Unsupervised Visual Representation Learning}} (Mar 2020)

\bibitem{heDeepResidualLearning2015}
He, K., Zhang, X., Ren, S., Sun, J.: Deep {{Residual Learning}} for {{Image Recognition}} (Dec 2015). \doi{10.48550/arXiv.1512.03385}

\bibitem{loshchilovDecoupledWeightDecay2019}
Loshchilov, I., Hutter, F.: Decoupled {{Weight Decay Regularization}} (Jan 2019). \doi{10.48550/arXiv.1711.05101}

\bibitem{oquabDINOv2LearningRobust2024}
Oquab, M., Darcet, T., Moutakanni, T., Vo, H., Szafraniec, M., Khalidov, V., Fernandez, P., Haziza, D., Massa, F., {El-Nouby}, A., Assran, M., Ballas, N., Galuba, W., Howes, R., Huang, P.Y., Li, S.W., Misra, I., Rabbat, M., Sharma, V., Synnaeve, G., Xu, H., Jegou, H., Mairal, J., Labatut, P., Joulin, A., Bojanowski, P.: {{DINOv2}}: {{Learning Robust Visual Features}} without {{Supervision}} (Feb 2024)

\bibitem{radfordLearningTransferableVisual2021}
Radford, A., Kim, J.W., Hallacy, C., Ramesh, A., Goh, G., Agarwal, S., Sastry, G., Askell, A., Mishkin, P., Clark, J., Krueger, G., Sutskever, I.: Learning {{Transferable Visual Models From Natural Language Supervision}} (Feb 2021). \doi{10.48550/arXiv.2103.00020}

\bibitem{raviSAM2Segment}
Ravi, N., Gabeur, V., Hu, Y.T., Hu, R., Ryali, C., Ma, T., Khedr, H., Rädle, R., Rolland, C., Gustafson, L., Mintun, E., Pan, J., Alwala, V., Carion, N., Wu, C.Y., Girshick, R., Dollár, P., Feichtenhofer, C.: {{SAM}} 2: {{Segment Anything}} in {{Images}} and {{Videos}}

\bibitem{sermanetTimeContrastiveNetworksSelfSupervised2018}
Sermanet, P., Lynch, C., Chebotar, Y., Hsu, J., Jang, E., Schaal, S., Levine, S.: Time-{{Contrastive Networks}}: {{Self-Supervised Learning}} from {{Video}} (Mar 2018). \doi{10.48550/arXiv.1704.06888}

\bibitem{sunExploringEffectiveFactors2023}
Sun, Y., Chen, Q., Wang, J., Wang, J., Li, Z.: Exploring {{Effective Factors}} for {{Improving Visual In-Context Learning}} (Apr 2023). \doi{10.48550/arXiv.2304.04748}

\bibitem{vanengelenSurveySemisupervisedLearning2020}
{van Engelen}, J.E., Hoos, H.H.: A survey on semi-supervised learning. Machine Learning  \textbf{109}(2),  373--440 (Feb 2020). \doi{10.1007/s10994-019-05855-6}

\bibitem{wangImagesSpeakImages2023a}
Wang, X., Wang, W., Cao, Y., Shen, C., Huang, T.: Images {{Speak}} in {{Images}}: {{A Generalist Painter}} for {{In-Context Visual Learning}} (Mar 2023). \doi{10.48550/arXiv.2212.02499}

\bibitem{wangMedCLIPContrastiveLearning2022}
Wang, Z., Wu, Z., Agarwal, D., Sun, J.: {{MedCLIP}}: {{Contrastive Learning}} from {{Unpaired Medical Images}} and {{Text}} (Oct 2022). \doi{10.48550/arXiv.2210.10163}

\bibitem{weissSurveyTransferLearning2016}
Weiss, K., Khoshgoftaar, T.M., Wang, D.: A survey of transfer learning. Journal of Big Data  \textbf{3}(1), ~9 (May 2016). \doi{10.1186/s40537-016-0043-6}

\bibitem{yuVisualTuning2024}
Yu, B.X., Chang, J., Wang, H., Liu, L., Wang, S., Wang, Z., Lin, J., Xie, L., Li, H., Lin, Z., Tian, Q., Chen, C.W.: Visual {{Tuning}}. ACM Computing Surveys  \textbf{56}(12),  1--38 (Dec 2024). \doi{10.1145/3657632}

\bibitem{zhangInstructMeMore2023}
Zhang, J., Wang, B., Li, L., Nakashima, Y., Nagahara, H.: Instruct {{Me More}}! {{Random Prompting}} for {{Visual In-Context Learning}} (Nov 2023). \doi{10.48550/arXiv.2311.03648}

\bibitem{zhangWhatMakesGood2023}
Zhang, Y., Zhou, K., Liu, Z.: What {{Makes Good Examples}} for {{Visual In-Context Learning}}? (Feb 2023). \doi{10.48550/arXiv.2301.13670}

\bibitem{zhuMedicalSAM22024}
Zhu, J., Hamdi, A., Qi, Y., Jin, Y., Wu, J.: Medical {{SAM}} 2: {{Segment}} medical images as video via {{Segment Anything Model}} 2 (Dec 2024). \doi{10.48550/arXiv.2408.00874}

\end{thebibliography}

\end{document}